%% file: main.tex
\documentclass[journal]{IEEEtai}
\IEEEoverridecommandlockouts
\usepackage{cite}
\usepackage{amsmath,amssymb,amsfonts}
\usepackage{algorithmic}
\usepackage{graphicx}
\usepackage{textcomp}
\usepackage[dvipsnames]{xcolor}
\usepackage{url}
\usepackage{romannum}
\def\BibTeX{{\rm B\kern-.05em{\sc i\kern-.025em b}\kern-.08em
    T\kern-.1667em\lower.7ex\hbox{E}\kern-.125emX}}

\newcommand{\paperTitle}{SNNLP: Energy-Efficient Natural Language Processing Using Spiking Neural Networks}

\newcommand{\paperKeywords}{Spiking Neural Networks, Low-Energy Computing, Natural Language Processing}

\begin{document}

\title{\paperTitle}

\author{R. Alexander Knipper, Kaniz Mishty, Mehdi Sadi, and Shubhra Kanti Karmaker Santu

\thanks{Manuscript received October 27, 2023; revised XXX XX, 202X. This work was supported in part by the National Science Foundation (NSF) under Grant Number CRII-2153394.}

\thanks{ The authors R. Alexander Knipper and Shubhra Kanti Karmaker Santu are with the Department of Computer Science and Software Engineering, Auburn University, Auburn, AL 36849, USA.
E-mail: rak0035@auburn.edu; sks0086@auburn.edu.}

\thanks{ The authors Kaniz Mishty and Mehdi Sadi are with the Department of Electrical and Computer Engineering, Auburn University, Auburn, AL 36849, USA.
E-mail: kzm0114@auburn.edu; mehdi.sadi@auburn.edu.}

}

\maketitle

\input{sections/abstract.tex}
\input{sections/impact.tex}


\input{sections/introduction.tex}
\input{sections/related.tex}
\input{sections/encoding.tex}

\input{sections/spiking.tex}

\input{sections/experiment_setup.tex}
\input{sections/results.tex}
\input{sections/inference.tex}

\input{sections/conclusion.tex}

\bibliographystyle{ieeetran}
\bibliography{anthology}

\input{biography}

\end{document}

%% file: sections/abstract.tex
\begin{abstract}

As spiking neural networks receive more attention, we look toward applications of this computing paradigm in fields other than computer vision and signal processing. One major field, underexplored in the neuromorphic setting, is Natural Language Processing (NLP), where most state-of-the-art solutions still heavily rely on resource-consuming and power-hungry traditional deep learning architectures. Therefore, it is compelling to design NLP models for neuromorphic architectures due to their low energy requirements, with the additional benefit of a more human-brain-like operating model for processing information. However, one of the biggest issues with bringing NLP to the neuromorphic setting is in properly encoding text into a spike train so that it can be seamlessly handled by both current and future SNN architectures.
In this paper, we compare various methods of encoding text as spikes and assess each method's performance in an associated SNN on a downstream NLP task, namely, sentiment analysis.
Furthermore, we go on to propose a new method of encoding text as spikes that outperforms a widely-used rate-coding technique, Poisson rate-coding, by around 13\% on our benchmark NLP tasks.
Subsequently, we demonstrate the energy efficiency of SNNs implemented in hardware for the sentiment analysis task compared to traditional deep neural networks, observing an energy efficiency increase of more than 32x during inference and 60x during training while incurring the expected energy-performance tradeoff.




\end{abstract}

%% file: sections/impact.tex
\begin{IEEEImpStatement}

As we see typical Natural Language Processing (NLP) architectures requiring vast amounts of computational power (e.g. ChatGPT, LLaMA, etc.), the issue of energy efficiency in NLP becomes more apparent than ever. Spiking neural networks (SNNs) offer vast energy savings under a new computing paradigm, which has its strengths in processing sequences. However, while most NLP technologies process language as a sequence, there is not a well-defined way to convert these text sequences into binary ``0/1'' spike trains for use in the energy-efficient paradigm introduced by SNNs.
In this paper, we recommend a method of encoding text as a spike train that outperforms other typically used methods by roughly 13\% on our benchmark NLP tasks while using less energy, allowing NLP to more readily be used in this energy-efficient environment.

\end{IEEEImpStatement}

%% file: sections/introduction.tex
\section{Introduction}

As we continue into the data-driven age, we find that more and more applications are leveraging machine learning techniques in some capacity. However, today's machine learning approaches bring up a glaring issue: these methods are incredibly energy-consuming, using countless floating-point operations to process large amounts of data. As we look toward implementing these machine learning (ML) applications in IoT devices, we need to reduce the energy consumption on learning tasks in order to increase the battery life of such devices.

Common approaches to reducing energy consumption to date mainly exist in improving the underlying hardware, while still adhering to the typical Von-Neumann architecture. However, a more promising approach lies in the \textit{Neuromorphic} architecture, where the goal is to create biologically-inspired computing chips for machine learning that deviates from the standard Von-Neumann approach \cite{neuromorphic-hardware}. This biologically-inspired approach places data and processing power in the same place—much like how the brain operates, reducing the overhead in moving data to/from a processor for intensive calculations like those done in typical ML approaches.

Using these biologically-inspired Spiking Neural Networks (SNNs), we can perform similar processing to typical Artificial Neural Networks (ANNs) using a network of neurons that exclusively communicate using binary ``0/1'' pulses. Furthermore, due to the nature of these `spiking neurons', spiking neural networks end up modeling the dependencies in input data over time, which aligns itself incredibly well with processing sequential data. With that in mind, natural language processing (NLP) presents itself as a promising candidate for implementations in spiking hardware, especially with the energy-intensive networks that currently dominate the state-of-the-art.

Current SNN research efforts focus mainly on two areas: computer vision and information processing~\cite{nunes2022spiking}. As a result of these efforts, spiking neural networks were proven both viable and useful in terms of achieving a comparable performance (with some expected performance degradation) to real-valued counterpart neural networks. Additionally, the spiking neural networks exhibit much lower energy consumption than their real-valued counterparts. However, many research areas, including NLP, are still underexplored in the spiking neural network setting and hence, it is compelling to study popular NLP tasks in this neuromorphic computing setting.
 

As text data is sequential in nature, NLP tasks in general lend themselves well to the processing paradigm of current spiking neural network architectures, which involve a time dimension and take in spike trains as inputs. However, one of the largest issues with bringing NLP to the neuromorphic setting is in properly encoding text into a spike train so it can be seamlessly handled by current and future SNN architectures.
Some typical methods that can be leveraged to do so include binarized word embeddings \cite{tissier2019near-lossless} and rate-coding techniques \cite{2015stdp}, but each method has its drawbacks.

In this paper, we explore the viability of processing text using spiking neural networks by applying them to a popular NLP problem: sentiment classification. In doing so, we aim to answer four major questions:
\begin{enumerate}
    \item Which works better for NLP tasks, directly using a binary embedding or rate-coding a floating-point embedding?
    \item If we make rate-coding a deterministic process as opposed to a stochastic one, do we see higher accuracy on NLP tasks?
    \item On NLP tasks, do we see the same energy-accuracy tradeoff that is reported in previous SNN literature?
    \item How far can we decrease the latency cost of SNNs while keeping performances competitive with traditional ANNs?
\end{enumerate}


\noindent Coupled with those major questions, we offer four major findings, as follows:
\begin{enumerate}
    \item While both binary embeddings and rate-coded floating-point embeddings perform similarly on downstream NLP tasks, directly using the binary word embedding offers slightly improved performance.
    \item We propose a method that makes rate-coding into a deterministic process, offering \textit{significantly} improved performance ($13\%$ increased accuracy) on downstream NLP tasks compared to a typical stochastic method, Poisson rate-coding.
    \item We observe a similar energy-performance tradeoff to results seen in previous SNN literature \cite{nunes2022spiking, ANN-SNN-conversion}, further grounding our results.
    \item Latency can be reduced to between $9.3\times$ and $17.5\times$ without significantly affecting accuracy, and can be reduced to between $3.7\times$ and $7\times$ when taking a performance hit of roughly $7$-$10\%$.
\end{enumerate}

\begin{figure}[!htb]
    \centering
    \vspace{-2mm}
   \includegraphics[scale = 0.39]{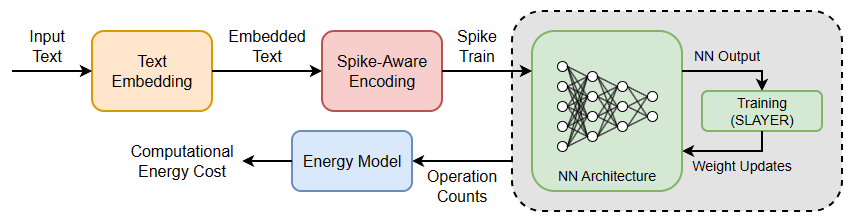}
    \vspace{-2mm}
    \caption{Overall encoding/training pipeline for both SNN and ANN. Text gets encoded into a continuous representation, converted into a spike train (for SNN usage), and sent to the downstream neural network for training/testing. During training/testing, operation counts are recorded so we can calculate appropriate energy numbers for comparison.}
    \label{fig:fig1}
\end{figure}

To be more specific, we show that, given an appropriate spike formatting for input text, emerging SNN training techniques are applicable to make inferences on text data. We then subsequently report the energy efficiency of SNN architectures applied to these tasks (Fig. \ref{fig:fig1}).
In summary, the paper is organized as follows:
Section \ref{sec:related} presents some background information and related work.
Section \ref{sec:encoding} explores different methods of encoding text for use in a spiking neural network.
Section \ref{sec:spiking} describes our methods and the energy model we use. 
Section \ref{sec:setup} describes our experimental setup, and 
section \ref{sec:results} presents our results on the sentiment classification problems, discussing our findings.
Finally, we conclude the paper in Section \ref{sec:conclusion}, discussing the limitations of our work and some possible future directions.

%% file: sections/related.tex
\section{Related Work}\label{sec:related}




\subsection{Spiking Neural Networks}
Spiking neural networks promise incredibly low-powered hardware \cite{neuromorphic-hardware}, adhering to a neuromorphic paradigm as opposed to the more energy-intensive Von-Neumann approaches we see in everyday commercial use. However, at present, spiking neural networks are difficult to train, lacking typical backpropagation methods and commonly-used activation functions. Furthermore, there is still a considerable gap in performance between a spiking neural network and its real-valued artificial neural network counterpart. To address this, three main approaches emerged, each aiming for a trained SNN with comparable performance to a similar ANN: 

\subsubsection{Conversion Approaches}

The most-studied approach has been the conversion approach, where an ANN is trained with backpropagation, and its learned weights are mapped to an SNN with the same architecture. So far, this approach has achieved the most consistent performance~\cite{ANN-SNN-conversion}.


\subsubsection{Unsupervised Learning Approaches}

Unsupervised learning in SNNs is mainly explored in applications or modifications of Spike Time-Dependent Plasticity (STDP) \cite{2015stdp}, and is perhaps the second most-studied approach for SNN training.

\subsubsection{Supervised Learning Approaches}

Arguably the most difficult approach for learning in SNNs is supervised learning, mostly due to the inability to calculate gradients from the spiking activation function. However, active research is proposing an alternate, ``surrogate" gradient approximation function. The most notable work in this area is SLAYER \cite{SLAYER}, which treats the spiking activations probabilistically, and approximates the gradient using the estimated probability density function of the spiking activation. In this work, we focus mainly on supervised learning and  utilize SLAYER for backpropagation during SNN training.

\subsection{Natural Language Processing}

\subsubsection{Text Encoding Methods}
In natural language processing, one of the key techniques applied in nearly every approach taken is to encode the input text into a continuous n-dimensional `feature vector' called an embedding~\cite{mikolov2013distributed}, mapping each token's (character, word, sentence, etc.) symbolic representation to a multidimensional latent vector space. One of the most important properties of this vector space is that it not only encodes the syntactic information of words but also encodes the proper semantic information, allowing related meanings to be extracted easily. Many word embedding techniques have been proposed so far, e.g., Word2Vec \cite{mikolov2013efficient}, GloVe \cite{pennington2014glove}, FastText \cite{bojanowski2017enriching}, etc.

Going beyond word embeddings, NLP researchers have also proposed methods that encode larger language constructs, such as phrases \cite{bojanowski2017enriching} or sentences \cite{conneau2017infersent, cer2018use}. These representations for larger language constructs have gained more traction in NLP research as the needs for more dense representations of text have become more apparent.

\subsubsection{Embedding Binarization}

While binarization efforts fall under efficient NLP, it is also a relatively underexplored field. However, one piece of notable work includes the proposal of a hash-based clustering technique for learning binary embeddings \cite{DBLP:journals/corr/JoulinGBDJM16}, where the authors concatenated the binary codes of the closest centroids for each word. Another notable method is to transform an existing real-valued embedding to a binary embedding using an auto-encoder~\cite{tissier2019near-lossless}, and yet another method is to learn correlations between one-hot encoded context and target blocks \cite{liang2021fruitfly}.
Furthermore, our own previous work explored a method of learning compressed binary embeddings from scratch using a genetic algorithm \cite{knipper-etal-2022-analogy}.

%% file: sections/encoding.tex
\section{Background on Spike-Aware Text Encoding}\label{sec:encoding}




In order to make proper use of a spiking neural network for natural language processing, we need to first encode our input text into a spike train. This can be done by converting already-existing feature vectors created for any text token (character, word, sentence, etc.) to binary spikes, either deterministically or non-determinstically.
In this work, we focus on comparing both deterministic and non-deterministic conversion approaches, assessing the accuracy and energy consumption of each.

\subsection{Word Embeddings}\label{sec:word_embed}

In the real-valued domain, word embeddings typically capture semantic and syntactic information by observing word co-occurrences and either predicting the target word given its context, predicting the context given a target word, or observing the ratios between global co-occurrences~\cite{mikolov2013distributed, pennington2014glove}.
To use a word embedding vector in an SNN, we need to convert said word vector from the real-valued domain to the spiking domain.
This involves utilizing either a binarization method or a rate-coding approach.

When using a binarization method, like the near-lossless binarization method proposed in \cite{tissier2019near-lossless}, one can convert a vector from a multi-hundred dimensional vector of floating-point numbers to a binary vector, comprised of "0/1" bits (commonly, embeddings are made up of 64, 128, 256, or 512 bits). This allows for a reduction in both the size and complexity of the word embeddings being used as input. While this method typically requires an autoencoder to generate the binary embeddings, this conversion is an operation that only needs to be done \emph{once}, so the energy cost incurred can be ignored under the assumption that the word embeddings being used were already converted in advance.

On the other hand, rate-coding is a well-known method of encoding floating-point information as a spike train \cite{2015stdp, 2016-rate-vs-temporal, 2020-deep-backprop, 2022-rate-vs-direct}. Using this method, floating-point numbers are easily converted to a series of spikes based on a Poisson process, where, for a number of timesteps $k$:

\input{equations/poisson.tex}

\noindent Given that $X \sim \text{Binomial}(k, p)$. This process is repeated for all items in the input vector $v$, generating a spike train for each of the input vector's feature dimensions.

In this work, we explore and compare both binarization and rate-coding approaches at the word level to gauge the reliability of each. Furthermore, we put forth an additional rate-coding technique that aims to increase the fidelity of the downstream SNN's input.



\subsection{Sentence Embeddings}


Unlike word embeddings, sentence embeddings typically utilize a Transformer \cite{vaswani2017transformer} model to encode sentences as a single vector representation, training to maximize the ability to seamlessly transfer sentence representations between different NLP tasks. With these representations, each sentence's relative meaning is also encoded, allowing these embeddings to be used in a similar manner to word embeddings, but at the sentence level. However, due to their reliance on the underlying Transformer architecture, sentence embeddings tend to inherently consume more energy than word embeddings in return for increased information density, information accuracy, and generalizability.

Similar to word embeddings, sentence embeddings can also be converted for use in an SNN via binarization techniques or rate-coding techniques. However, it is important to know that the majority of sentence embeddings generate their sentence embeddings \emph{each time} a new dataset is introduced, as opposed to only being calculated once. This makes sentence embeddings far more expensive energy-wise compared to word embeddings. As such, we make use of sentence embeddings in this work purely to help illustrate the potential performance of spike-aware text encoding approaches.

In this work, we make use of Universal Sentence Encoder (USE) \cite{cer2018use} to encode each of our input sentences into a single 512-dimension floating-point vector, allowing for a compact representation of each text sample. Once again, in order to use these embedded representations in an SNN, we need to convert this floating-point vector into a binary vector using either a binarization method or a rate-coding method. In this work, we opt to utilize various rate-coding approaches, as binarization methods typically involve an autoencoder \emph{in addition to} the Transformer model being used to generate the embedding in the first place, which is far too expensive energy-wise.

%% file: equations/poisson.tex
\begin{equation}
Y_i = \begin{cases}
1 & \text{if } X_i < p_i \\
0 & \text{if } X_i \geq p_i
\end{cases}
\label{eq:poisson}
\end{equation}

%% file: sections/spiking.tex
\section{Sentiment/Emotion Classification using SNN}\label{sec:spiking}




To demonstrate an SNN's effectiveness in performing natural language processing tasks, we applied it to a simple, widely-used task called sentiment classification.  By extension, we also demonstrate various possible methods of preprocessing text data for use in an SNN. As described in Section \ref{sec:setup}, we choose to test the SNN on both a 2-class sentiment classification task and a 6-class emotion classification task to demonstrate its effectiveness and benchmark its performance.

\subsection{Network Type}
While sentiment classification tasks are currently solvable with a high accuracy ($>94\%$) using state-of-the-art neural architectures like Transformers \cite{vaswani2017transformer}, SNNs do not yet have these complex architectures reliably available to them. Therefore, in order to perform an apple-to-apple comparison, we opt to implement both the spiking and real-valued neural network architectures using a fully-connected, feedforward neural network on both of these tasks, and subsequently, report each downstream SNN's accuracy and energy consumption compared to the equivalent, real-valued ``shadow'' network.

\subsection{Encoding Methods}


We test various encoding methods in order to find the best way to properly utilize the time-based property of the network. Namely, we utilize the following method-model combinations, both at the sentence and word-level (with the exception of SNN-bin, which is only used at the word-level):

\begin{itemize}
    \item \textbf{ANN} - This is a typical ANN using regular floating-point embeddings
    \item \textbf{SNN-rate} - This is a custom, deterministically rate-coded embedding being used in a downstream SNN
    \item \textbf{SNN-rate-rand} - This is a Poisson rate-coded embedding being used in a downstream SNN
    \item \textbf{SNN-bin} - This is a binarized embedding being used in a downstream SNN
\end{itemize}

For our comparisons, we use an inference window of 50ms with the exception of SNN-bin, which instead takes in a fixed-length context of 512 for the 2-class task and 32 for the 6-class task. In this method, the input text is encoded at the word-level and formatted into spike trains as-is, so a sufficiently-long context is required to represent all of the words in a given input sequence.

Our other methods (ANN, SNN-rate, and SNN-rate-rand) leverage the average of all word embeddings in the input sequence, which allows us to get an approximation of a sentence-level embedding without the associated energy cost. Once we have this averaged embedding, we then encode it with our rate-coding methods for use in our downstream SNNs. As explained in section \ref{sec:word_embed}, the Poisson rate-coding process is a rather popular encoding method, and results in a stochastic set of spike trains that roughly approximate the floating-point number that was encoded. However, this approach raises a question: What if we deterministically encode the spikes such that the floating-point value being encoded is represented with higher fidelity? Would this higher-fidelity spiking rate result in a corresponding higher performance?

Our custom rate-coding method, which we refer to as ``SNN-rate'' in this work, aims to address that question by modifying the inherently stochastic Poisson rate-coding method (referred to as ``SNN-rate-rand'') to instead generate deterministic spike trains for the given embedding. We generate our deterministic encoding as follows:

\input{equations/modified_lif.tex}

Where $x_i$ is the floating-point value being encoded, $V_i(t)$ is the membrane potential of the neuron at timestep $t$, and $\alpha_i$ is the membrane threshold of the neuron.
Essentially, we modify the standard Leaky Integrate-and-Fire (LIF) neuron model in the following ways:
\begin{itemize}
    \item Make the neuron take in a floating-point value: the self-accumulation parameter
    \item Every timestep, perform the accumulation operation using \emph{only} the value stored as the self-accumulation parameter, which we refer to as the self-accumulation operation
\end{itemize}
\noindent It is important to note that this modified ``Self-Accumulate-and-Fire" (SAF) neuron model (Fig. 2) only takes as input the self-accumulation parameter, $x_i$, with the clock, $t$, telling the neuron how many times to perform the self-accumulation operation. Furthermore, when the neuron receives a new floating-point input, that input value becomes the neuron's new self-accumulation parameter, $x_i$, for all future self-accumulation operations.

\begin{figure}[!htb]
    \centering
    \vspace{-2mm}
   \includegraphics[scale = 0.57]{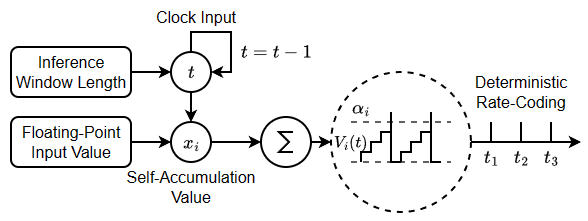}
    \vspace{-2mm}
    \caption{Illustrated ``Self-Accumulate-and-Fire" (SAF) neuron. Takes in a floating-point value and self-accumulates it over time, spiking similarly to a standard LIF neuron.}
    \label{fig:saf_neuron}
\end{figure}

The SAF neuron's output is then recorded as our rate-coded spike train and is sent directly to the rest of the SNN. This modified ``Self-Accumulate-and-Fire" (SAF) neuron, designed to enhance the integration of rate-coding behavior into the SNN model's architecture, facilitates our deterministic rate-coding process. This approach lends itself well to the spiking paradigm, particularly when formulated as a modified neuron model.



\subsection{Training Method}\label{sec:training}


While using the various spike-aware text encodings discussed in Section \ref{sec:encoding}, we opt to train our downstream SNNs using SLAYER \cite{SLAYER}, which allows the SNN to update its weights during training using a surrogate gradient for backpropagation. We additionally train two ``baseline'' ANNs for each task: 1) A regular ANN using the floating-point sentence embeddings, and 2) An ANN using floating-point word embeddings.

For all networks, we maintain their network architectures to be as identical as possible. All networks utilize an input size eqivalent to the size of the embedding technique being used. The input sizes used for each embedding method are as follows:

\begin{itemize}
    \item Floating-point \& rate-coded sentence embeddings: $512$
    \item Floating-point \& rate-coded word embeddings: $200$
    \item Binarized word embeddings: $64$
\end{itemize}

\noindent All networks have two hidden layers, sized at 256 and 128 neurons respectively. Finally, we size the output layer to match the number of output classes in each benchmark task, i.e. 2 neurons for sentiment classification and 6 neurons for emotion classification.


All networks use a learning rate ($\eta$) of $1e^{-5}$, and train for 50 epochs. The ANNs use cross-entropy loss as their loss function, while the SNNs use the spike rate as their loss function. Namely, the SNNs aim for the correct output class' neuron to spike for a majority of the inference window and all other output class' neurons to spike far less.

Furthermore, the SNNs all use the following neuron parameters:
\begin{itemize}
    \item threshold: $1.25$
    \item current\_decay: $0.25$
    \item voltage\_decay: $0.03$
    \item tau\_grad: $0.03$
    \item scale\_grad: $3$
\end{itemize}
\noindent Where the threshold and decay values determine the general behavior of the Leaky Integrate and Fire (LIF) neuron, that is, when the neuron fires and how much membrane potential is going to decay off per timestep. Additionally, the tau\_grad and scale\_grad values adjust the time constant and scale of the gradient, respectively.

\subsection{Energy Model}


The energy efficiency of an SNN comes from two main features. First, there is less computational demand due to the event-driven binary activity of the neurons. In an SNN, a computation is performed only if an incoming spike is received. The neuron's membrane potential $V$ at time $t$ can be expressed as $V(t)=V(t-1)+L+\sum_{i=1}^{N} w_i \sum_{k=1}^{\infty} \delta_i(t-k)$ \cite{neuromorphic-hardware,ANN-SNN-conversion}. The parameter  $w_i$ represents the weight of synapse $i$, $\delta$ is the delta function, $L$ is the leak term and $N$ is the number of synaptic weights. This equation implies that in the presence of a spike at time-stamp $t$, the synaptic weight is simply added with the existing membrane potential.   These computations can be mapped as Accumulation (AC) operations and performed by an Adder/Accumulator circuit. On the other hand, a typical RNN/ANN neuron requires one/two Multiplication and one Accumulation (MAC) operation. Second, the number of both on and off-chip memory accesses is much smaller for an SNN compared to an ANN. In a regular ANN, most of the on-chip memory is occupied by the large intermediate activation; sometimes, we even need to go to the off-chip to store the internal states. As a result, there are more off-chip accesses to fetch the network parameters and activations. In addition, ANNs require frequent on-chip accesses (depending on the size of the computation array unit and on-chip memory size) to store the partial sums of the output activations. However, in an SNN during inference, we only need to fetch the synaptic weights and initial inputs from off-chip once.

To model the total system energy consumed by an SNN in operation, we dissect the total energy into two parts: (\romannum{1}) energy associated with the computation and (\romannum{2}) energy associated with the memory access. This paper only models the energy required for the computations (case \romannum{1}), and modeling the system energy is considered future work as it involves detailed modeling of the hardware's memory system hierarchy.

We count the number of operations per layer per sample (for each NN) to estimate the computation energy of the SNN and ANN networks \cite{yin2020effective}. An ANN layer $l$ with $N$ neurons will require ($N_{l-1}N_{l}$) MAC operations to produce its output. If $E_{MAC}$ is the energy cost of a single MAC operation, the total cumulative energy consumed ($ E_{ANN_{total}}$) in the computation of $L$ layers is modeled as follows:

\vspace{-2mm}
\begin{equation}
    E_{ANN_{total}} = \sum_{l=1}^{L} (N_{l-1}N_{l})E_{MAC}
    \label{eq:ann_enrg}
\end{equation}
A spiking neuron on specialized SNN hardware will perform an Accumulate (AC) operation per synapse if a spike is received. An SNN layer $l$ with $N$ neurons upon the reception of a spike needs to perform $N_{l-1}N_{l}$ AC operations. However, depending on the fraction of neuron firing rate, $S$, not all the neurons will remain perpetually active \cite{yin2020effective}. Therefore, the total energy consumed in the computation of $L$ layers over $K$ timesteps is modeled as:

\vspace{-2mm}
\begin{equation}
    E_{SNN_{total}} = \sum_{t=1}^{K} \sum _{l=1}^{L}(N_{l-1}S_{l-1}N_{l}) E_{AC}
    \label{eq:ssn_enrg}
\end{equation}

$S$ is always $<<$1, making SNNs computation-efficient compared to ANNs. 
In addition, an AC operation consumes less than half of the energy a MAC operation consumes, as shown in Table \ref{tab:mac_add_pow}.

%% file: equations/modified_lif.tex
\begin{equation}
\emph{$V_i(t) = V_i(t-1) + x_i(t)$} 
\label{eq:modified_lif}
\end{equation}
\begin{equation}
Y_i(t) = \begin{cases}
1 & \text{if } V_i(t) \geq \alpha_i \\
0 & \text{otherwise }
\end{cases}
\label{eq:modified_lif_firing}
\end{equation}

%% file: sections/experiment_setup.tex
\section{Experimental Setup}\label{sec:setup}

In this section, we describe how we evaluate and compare each implemented SNN and its counterpart real-valued ANN.

\subsection{Dataset}

For our experiments, we used the IMDb Movie Review dataset \cite{imdb_sentiment} for our simple 2-class sentiment classification problem and the CARER dataset \cite{saravia-etal-2018-carer} for our multiclass (6 labels) emotion classification problem. The datasets are split into training, validation, and testing sets as indicated in Table \ref{tbl:dataset}, and the tasks associated with each dataset are briefly described below:

\begin{itemize}
    \item IMDb Movie Review: Given a string of text from a movie review, decide whether the text is expressing positive or negative sentiment.

    \item CARER: Given a tweet as text, decide which of the 6 emotions - sadness, joy, love, anger, fear, or surprise - are being exhibited in the text.
\end{itemize}

\input{tables/dataset.tex}

\subsection{Implementation Details}

\begin{figure}[ht]
    \centering
    \includegraphics[scale = 0.345]{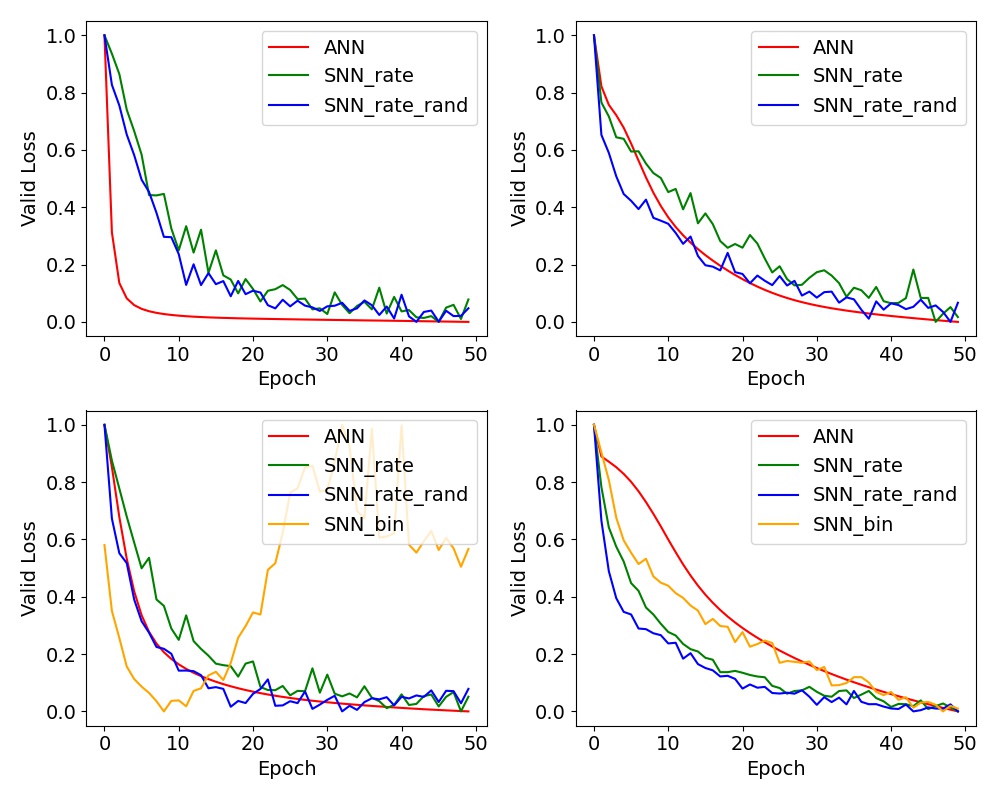}
    \caption{All model training curves. The top two graphs depict curves for sentence-level methods and the bottom two graphs show word-level methods. The left-side graphs show models trained for the IMDB 2-class task, where the right-side graphs show models trained for the 6-class emotion classification task.}
    \label{fig:model_training}
\end{figure}

For all models, we test the same model architecture on both problems, using the layer sizes described in Section \ref{sec:training}.
We train all models for 50 epochs, recording their convergence in Figure \ref{fig:model_training}.

\subsection{Evaluation Metrics}

When evaluating our models for comparison, we report two main performance metrics: inference accuracy and the mean reciprocal rank (MRR). We report accuracy numbers for both tasks, but only report the MRR for the emotion classification task, as it having more than two output classes means we can benefit from knowing where the desired class sits in the overall predicted ranking.

With the MRR measurements, we aim to examine how close the correct class was to the top spot if we had ranked the output neurons in terms of how many times they actually fired. For a given sample set, we define MRR as:

\input{equations/mrr.tex}

\noindent where $S$ is the collection of testing samples and $rank_i$ indicates the ordinal position of the correct class label for a particular sample $i$.

%% file: tables/dataset.tex
\begin{table}[]
    \centering
    \caption{Number of training, validation, and inference examples used for each evaluation dataset}
    \label{tbl:dataset}
    \begin{tabular}{c||c|c|c|c}
        \hline
         Dataset    & Training & Validation & Inference & Total \\
         \hline\hline
         IMDb       & 30,000   & 5,000      & 15,000    & 50,000 \\
         \hline
         CARER      & 12,000   & 2,000      & 6,000     & 20,000 \\
         \hline
    \end{tabular}
    
\end{table}

%% file: equations/mrr.tex
\begin{equation}
\emph{$MRR = \frac{1}{|S|} \sum_{i=1}^{|S|} \frac{1}{rank_i}$} 
\label{eq:mrr}
\end{equation}

%% file: sections/results.tex
\section{Experimental Results and Analysis}\label{sec:results}



\subsection{Model Performance}


In this section, we focus on answering our first two questions: 1) Which works better for NLP, binary embeddings or rate-coded floating-point embeddings? 2) If we make rate-coding a deterministic process as opposed to a stochastic one, do we see a performance improvement?
In addressing these questions, we test our SNN on two NLP tasks:
\begin{itemize}
    \item IMDb Review dataset: 2-class sentiment classification
    \item CARER dataset: 6-class emotion classification
\end{itemize}

We outline our results on both tasks in Tables \ref{tbl:sentiment_performance}, \ref{tbl:emotion_performance}, \ref{tbl:sentiment_word_performance}, and \ref{tbl:emotion_word_performance}. Note that we do not record the mean reciprocal rank (MRR) for the 2-class sentiment classification problem, as it is not meaningful for a binary classification task.

As shown in Tables \ref{tbl:sentiment_performance}, \ref{tbl:emotion_performance}, \ref{tbl:sentiment_word_performance}, and \ref{tbl:emotion_word_performance}, we demonstrate the expected performance gap between the ANN and our SNNs. Furthermore, both the word-level and sentence-level models are largely competitive with each other, with the sentence-level models performing better on 2-class classification (Table \ref{tbl:sentiment_performance}) and word-level models performing better with multi-class classification (Table \ref{tbl:emotion_word_performance}).

Additionally, it appears that at the word-level, we see similar performances from Poisson rate-coded spike trains and spike trains generated directly from a binary word embedding (Tables \ref{tbl:sentiment_word_performance} and \ref{tbl:emotion_word_performance}). In all cases, we observe slightly improved performance when using a binary word embedding as opposed to using a Poisson rate-coded spike train. However, we see a significant performance increase across all measures, embeddings, and tasks when the rate-coded spike trains are generated deterministically. This also exhibits a similarly significant improvement over using the binary embedding directly.

It is important to note that most state-of-the-art results on these tasks are achieved using Transformer architectures, like BERT \cite{devlin2018bert} and XL-NET \cite{yang2019xlnet}, which are more complex (and energy-intensive) models than fully-connected neural networks. Such architectures are currently being researched, and stable implementations are not yet available for neuromorphic hardware.

Overall, our results demonstrate the following conclusions:
\begin{itemize}
    \item We observe the expected performance degradation going from an ANN to its SNN counterpart, which is justified by the energy improvements explored in Section \ref{sub_sec:energy}.
    \item We observe a similar performance between Poisson rate-coded and binary embedding spike trains, with the binary embedding being slightly more effective than typical Poisson rate-coding in both tasks.
    \item We observe a significant increase in performance when using deterministically-encoded spike trains as opposed to Poisson rate-coded spike trains. By extension, we see a similarly significant improvement using the deterministically-encoded spike trains over the binary embedding-based spike trains as well.
\end{itemize}




\input{tables/sentiment_sentence_performance.tex}

\input{tables/emotion_sentence_performance.tex}

\input{tables/sentiment_word_performance.tex}

\input{tables/emotion_word_performance.tex}

\subsection{Energy Efficiency}
\label{sub_sec:energy}

In this subsection, we demonstrate the energy improvement of our implemented word-level, SNN-based architecture over an equivalent, conventional ANN for performing sentiment and emotion classification on the IMDb Movie Review and CARER datasets, respectively. We design an FP32 Multiplier-Accumulator unit and an FP32 Accumulator unit to carry out the GEMM operations of ANNs and the accumulation operations of SNN tasks. We synthesize and perform PNR (place and route) of the design using Synopsys 14-nm Standard Cell Library \cite{saed14} at a 1GHz clock frequency. Each unit's energy consumption per operation is listed below in Table \ref{tab:mac_add_pow}.

\input{tables/add_mac_pwr.tex}

\begin{figure}[ht]
    \centering
    \includegraphics[scale = 0.60]{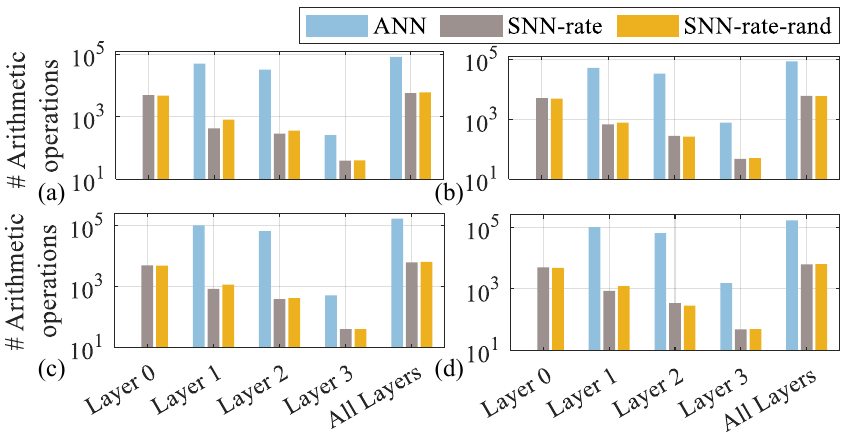}
    \caption{SNN vs ANN number of arithmetic operations of \emph{Sentiment Classification} during inference (a) \& training (c) and \emph{Emotion Classification} during inference (b) \& training (d).}
    \label{fig:snn_ann_op_count}
\end{figure}

\begin{figure}[ht]
    \centering
    \includegraphics[scale = 0.62]{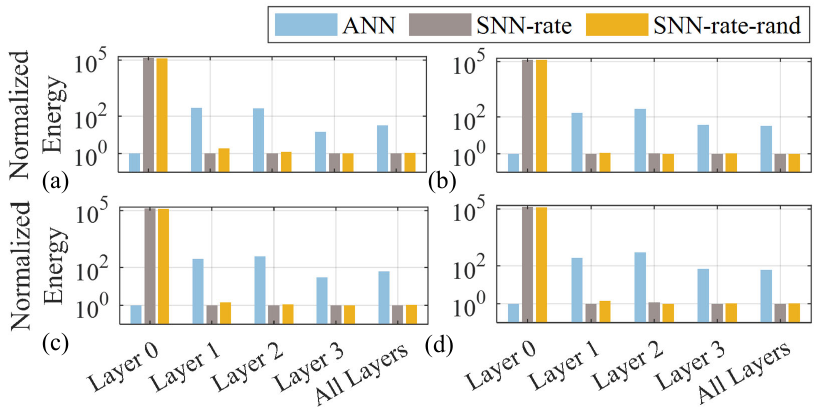}
    \caption{Improvement of SNN in computation energy over ANN. \emph{Sentiment Classification} during inference (a) and training (c). \emph{Emotion Classification} during inference (b) and training (d).}
    \label{fig:snn_ann_pwr}
\end{figure}

In Fig. \ref{fig:snn_ann_op_count}, we plot the layer-wise and total computations per sample for the ANN and SNN architectures while performing the sentiment classification ((a)-inference, ((c)-training) and emotion classification ((b)-inference, (d)-training) tasks. As shown, ANNs appear to require an order of magnitude more operations than SNNs. Overall, the total number of computations in ANNs for sentiment and emotion classification are $\sim 15 \times$  of SNNs during \textit{inference} (Fig. \ref{fig:snn_ann_op_count} (a)-(b)).
As the SNNs run for 50 timesteps per sample (equating roughly 50ms) during inference, and this ``inference window'' can be adjusted, this value will increase/decrease along with the configured inference window size. 
Additionally, to compare \textit{training} costs on a per-epoch basis, we averaged the number of computations per sample over all 50 epochs, which is reported in Fig.~\ref{fig:snn_ann_op_count} (c)-(d). As shown, ANNs require even more operations to train their parameters: $\sim 27 \times$ more than SNNs. Furthermore, we again emphasize that MAC operations dominate the ANNs while SNNs are dominated by AC operations, further increasing our energy improvements.

To showcase this improvement, we estimate the energy consumption of the ANN and SNN for our two use cases using equations \ref{eq:ann_enrg} \& \ref{eq:ssn_enrg}, and the energy values from Table \ref{tab:mac_add_pow}. As shown in Fig. \ref{fig:snn_ann_pwr}, the SNN achieves roughly $32\times$ more energy efficiency during inference (Fig. \ref{fig:snn_ann_pwr} (a)-(b)) and roughly $60\times$ during training (Fig. \ref{fig:snn_ann_pwr} (c)-(d)) compared to the ANN for sentiment and emotion classification tasks.
However, we do observe calculations taking place in the SNNs' input layer (\textit{Layer 0}), whereas the ANNs' input layers do not have any extra calculations.
This is due to the input layer being comprised of spiking neurons - since the input is a spike train, it will naturally follow the Leaky Integrate-and-Fire (LIF) behavior, which incurs extra operations.
Nonetheless, despite the extra layers' worth of operations and energy consumption, SNNs still achieve exceptional energy savings compared to ANNs - roughly $32\times$ during inference and $60\times$ during training.

Overall, this means that our SNN-NLP architectures achieve the expected energy improvements in return for the previously observed performance loss incurred.

\subsection{Latency Impact}
\vspace{-0.0in}
Furthermore, we measure the amount of time taken to perform inference on both the ANN and the SNN to determine the effect on the per-sample inference latency. In its current configuration, the SNNs take $50ms$ to infer each sample for both tasks, although that is a value that can be adjusted, as explored in section \ref{sec:inference}. The ANNs, on the other hand, take about $2.69ms$ per sample for the sentiment classification task and $1.429ms$ per sample for the emotion classification task. This means the SNNs have an approximate $18.53\times$ increase in inference latency over the associated ANN for 2-class sentiment classification and a $34.99\times$ increase for multi-class emotion classification. Overall, our observed latency increase for SNN-NLP conforms with previously-reported SNN latency increases over regular neural networks for image recognition tasks \cite{ANN-SNN-conversion}. Furthermore, this latency increase is justified for low-power applications in edge computing.


%% file: tables/sentiment_sentence_performance.tex
\begin{table}[]
    \centering
    \caption{Sentence-level 2-class sentiment classification:\\ Accuracy}
    \label{tbl:sentiment_performance}
    \begin{tabular}{c||c|c}
        \hline
         Model
         & \begin{tabular}[c]{@{}c@{}}Training\\ Accuracy\end{tabular}
         & \begin{tabular}[c]{@{}c@{}}Inference\\ Accuracy\end{tabular}   \\
         \hline\hline
         ANN            & 0.859     & 0.867     \\
         \hline
         SNN-rate       & \textbf{0.836}     & \textbf{0.840}     \\
         \hline
         SNN-rate-rand  & 0.659     & 0.656     \\
         \hline
    \end{tabular}
    
\end{table}

%% file: tables/emotion_sentence_performance.tex
\begin{table}[]
    \centering
    \caption{Sentence-level 6-class emotion classification:\\ Accuracy \& MRR}
    \label{tbl:emotion_performance}
    \begin{tabular}{c||c|c|c}
        \hline
         Model
         & \begin{tabular}[c]{@{}c@{}}Training\\ Accuracy\end{tabular}
         & \begin{tabular}[c]{@{}c@{}}Inference\\ Accuracy\end{tabular}
         & \begin{tabular}[c]{@{}c@{}}Inference\\ MRR\end{tabular}   \\
         \hline\hline
         ANN            & 0.617     & 0.633     & 0.778     \\
         \hline
         SNN-rate       & \textbf{0.534}     & \textbf{0.522}     & \textbf{0.703}     \\
         \hline
         SNN-rate-rand  & 0.410     & 0.395     & 0.609     \\
         \hline
    \end{tabular}
    
\end{table}

%% file: tables/sentiment_word_performance.tex
\begin{table}[]
    \centering
    \caption{Word-level 2-class sentiment classification:\\ Accuracy}
    \label{tbl:sentiment_word_performance}
    \begin{tabular}{c||c|c}
        \hline
         Model
         & \begin{tabular}[c]{@{}c@{}}Training\\ Accuracy\end{tabular}
         & \begin{tabular}[c]{@{}c@{}}Inference\\ Accuracy\end{tabular}   \\
         \hline\hline
         ANN            & 0.821     & 0.824     \\
         \hline
         SNN-rate       & \textbf{0.788}     & \textbf{0.788}     \\
         \hline
         SNN-rate-rand  & 0.628     & 0.631     \\
         \hline
         SNN-bin        & 0.650     & 0.664     \\
         \hline
    \end{tabular}
    
\end{table}

%% file: tables/emotion_word_performance.tex
\begin{table}[]
    \centering
    \caption{Word-level 6-class emotion classification:\\ Accuracy \& MRR}
    \label{tbl:emotion_word_performance}
    \begin{tabular}{c||c|c|c}
        \hline
         Model
         & \begin{tabular}[c]{@{}c@{}}Training\\ Accuracy\end{tabular}
         & \begin{tabular}[c]{@{}c@{}}Inference\\ Accuracy\end{tabular}
         & \begin{tabular}[c]{@{}c@{}}Inference\\ MRR\end{tabular}   \\
         \hline\hline
         ANN            & 0.574     & 0.572     & 0.734     \\
         \hline
         SNN-rate       & \textbf{0.542}     & \textbf{0.536}     & \textbf{0.711}     \\
         \hline
         SNN-rate-rand  & 0.424     & 0.432     & 0.633     \\
         \hline
         SNN-bin        & 0.446     & 0.432     & 0.635     \\
         \hline
    \end{tabular}
    
\end{table}

%% file: tables/add_mac_pwr.tex


\begin{table}[]
    \centering
    \caption{FP MAC and ACC module energy}
    \label{tab:mac_add_pow}
    \begin{tabular}{c||c|c} \hline
       Module  & Tech. (nm) & Energy (pj)\\ \hline \hline
        FP Multiplier-Accumulator (FP-MAC) & 14 & 57.14 \\ \hline
        FP Accumulator (FP-ACC) & 14 & 25.63 \\ \hline
    \end{tabular}    
    
\end{table}

%% file: sections/inference.tex
\section{Effects of Inference Window Size}\label{sec:inference}

In an SNN, we can adjust how long the model runs for, that is, the \emph{inference window} of the model. In essence, it dictates how many timesteps' worth of spikes are sent to the model as input, as well as how many timesteps' worth of output spikes the model can make. As common sense would dictate, making the inference window larger would result in a performance increase at the cost of increased latency and energy consumption. Conversely, decreasing the inference window would result in a performance decrease with an associated reduction in both latency and energy consumption. In this section, we aim to both confirm that common sense assumption \emph{and} find an acceptable performance/latency tradeoff for our downstream SNN models.


\begin{figure}[ht]
    \centering
    \includegraphics[scale = 0.34]{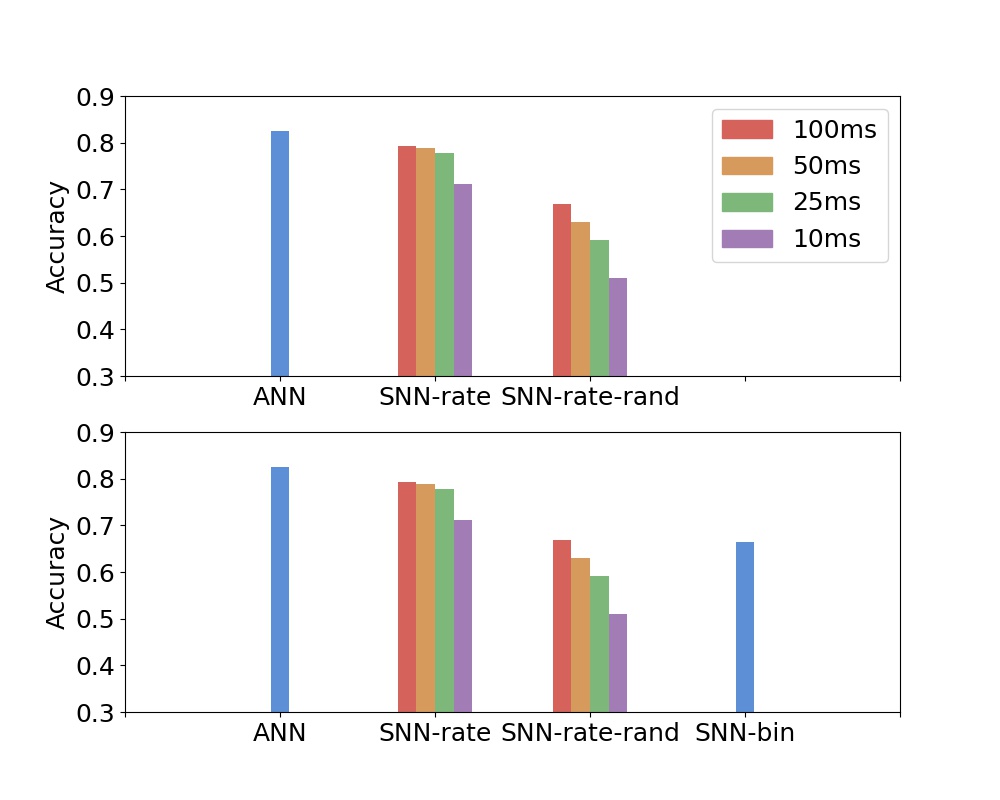}
    \caption{Accuracy based on inference window size for the IMDb 2-class sentiment classification task. The top graph depicts performance at the sentence-level and the bottom graph depicts performances at the word-level.}
    \label{fig:imdb_window_accuracy}
\end{figure}

\begin{figure}[ht]
    \centering
    \includegraphics[scale = 0.34]{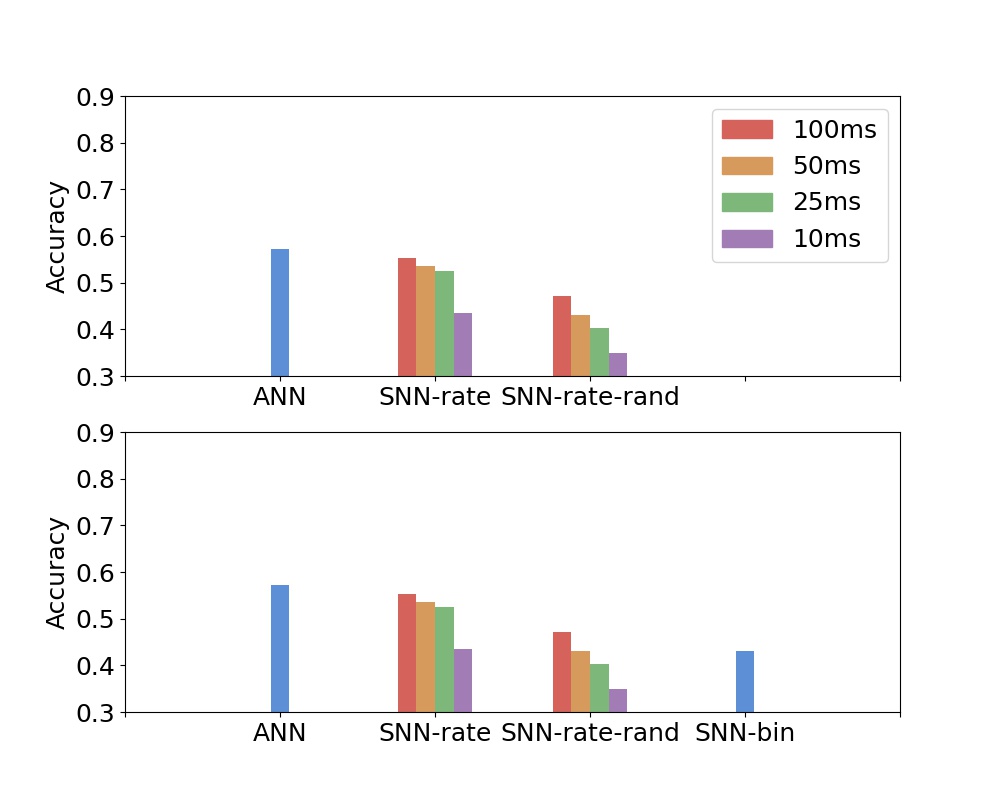}
    \caption{Accuracy based on inference window size for the Twitter Emotion 6-class sentiment classification task. The top graph depicts performance at the sentence-level and the bottom graph depicts performances at the word-level.}
    \label{fig:emotion_window_accuracy}
\end{figure}

\begin{figure}[ht]
    \centering
    \includegraphics[scale = 0.34]{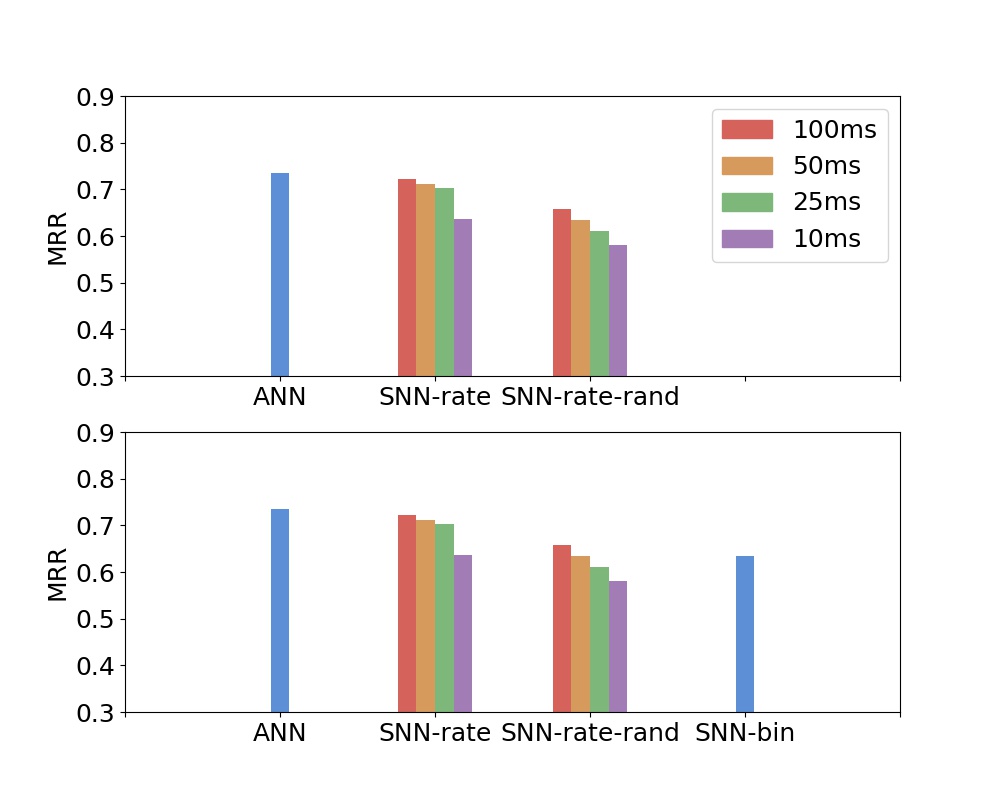}
    \caption{Mean Reciprocal Rank (MRR) based on inference window size for the Twitter Emotion 6-class sentiment classification task. The top graph depicts performance at the sentence-level and the bottom graph depicts performances at the word-level.}
    \label{fig:emotion_window_mrr}
\end{figure}

For our experiments, we vary the inference window between the following values: 100ms, 50ms (The inference window used in all other results), 25ms, and 10ms, where each millisecond is counted as a single timestep. Our results are shown in Fig. \ref{fig:imdb_window_accuracy}, \ref{fig:emotion_window_accuracy}, and \ref{fig:emotion_window_mrr}. As shown, we see the expected performance degradation as we reduce the size of the inference window, with performance dropping off more sharply between 25ms and 10ms. As a result, we recommend that the inference window for most downstream models stays between 10 and 25ms, as this achieves a performance-latency tradeoff where SNNs remain competitive with ANNs ($\sim 7$-$10\%$ less accuracy), and latency drops to $\sim 3.7$-$7\times$ that of equivalent ANNs.

%% file: sections/conclusion.tex
\section{Conclusions \& Future Directions}\label{sec:conclusion}

In this work, we explore the viability of spiking neural networks for performing natural language processing tasks, describing various encoding methods and testing various SNNs on two downstream (sentiment/emotion) classification tasks. Overall, we present the following major contributions:
\begin{itemize}
    \item At the word-level, we demonstrate that both Poisson rate-coding and directly using the binary embedding perform similarly.
    \item By making rate-coding a deterministic process, we see a significant improvement in accuracy over Poisson rate-coding (roughly $13\%$). This narrows the performance gap between the SNNs and ANNs to roughly $3.6\%$ at the word-level, further underscoring SNNs' viability on NLP tasks.
    \item We observe the expected energy-accuracy tradeoff between the ANN and the SNN, as seen in previous SNN literature \cite{nunes2022spiking, ANN-SNN-conversion}, further grounding our results.
    \item We successfully reduce the SNNs' latency to $9.3-17.5x$ that of the ANNs without significantly affecting performance, and further reduce the latency to $3.7-7x$ when taking an additional performance hit or roughly $7-10\%$.
\end{itemize}
\noindent These results show that applying SNNs to NLP tasks is not only viable, but it can be \emph{useful}, especially in limited-resource environments where more energy-efficient methods are required. Additionally, we demonstrate that the SNNs are extremely power-efficient, achieving an approximate $32\times$ and $60\times$ energy savings during inference and training, respectively. Furthermore, this energy efficiency is achieved \emph{while} narrowing the performance gap between ANNs and SNNs to under 5\% accuracy.



\subsection{Limitations}
This work is limited in a few ways, which we aim to address as we move forward with this direction. The main limitations of this work are as follows:
\begin{itemize}
    \item We mainly test our SNNs on text classification tasks; more research is needed to verify the generalizability of SNNs to other, more complex NLP tasks (e.g., language translation) and architectures (e.g., Transformer, BERT).
    \item While for NLP tasks the SNNs result in a vast increase in energy efficiency, this comes at a slight cost to both model performance and inference latency. However, this energy-latency trade-off is well-known for SNNs \cite{nunes2022spiking, ANN-SNN-conversion}.
\end{itemize}

\subsection{Future Directions}

For further work, we plan to \emph{fully} leverage the time-series nature of SNNs and test its ability to process arbitrarily long sequences of text using a series of embedded data. To that end, we also plan to test SNNs on a wider variety of tasks to further cement its effectiveness and usefulness, and test more methods of encoding text into spikes in an energy-efficient manner. Finally, we plan to incorporate the memory-access-related energy in our energy model and evaluate the system-level energy improvements of SNNs over ANNs.

\subsection{Reproducibility}

All code, data, and experiments contained in this paper are available on GitHub at \underline{https://github.com/alexknipper/SNNLP}.


%% file: biography.tex
\begin{IEEEbiography}[{\includegraphics[width=\linewidth]{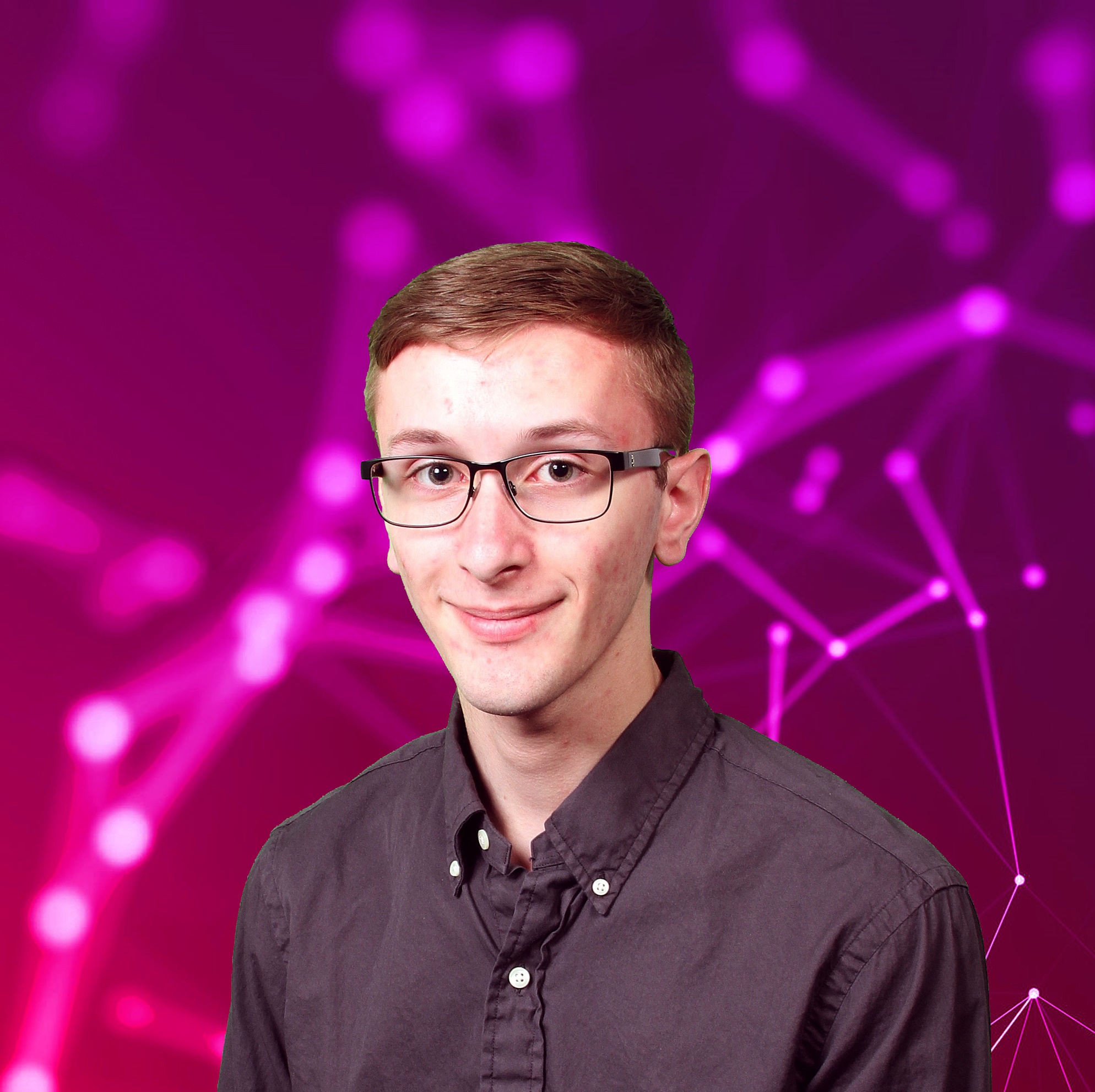}}]{R. Alexander Knipper}
is a PhD student in the Department of Computer Science and Software Engineering at Auburn University, Auburn, AL. He received a BS degree from Indiana Institute of Technology in Fort Wayne, IN in 2020. His current research work is focused on developing conversational systems and working with Natural Language Processing in Neuromorphic architectures.
\end{IEEEbiography}

\begin{IEEEbiography}[{\includegraphics[width=1\linewidth]{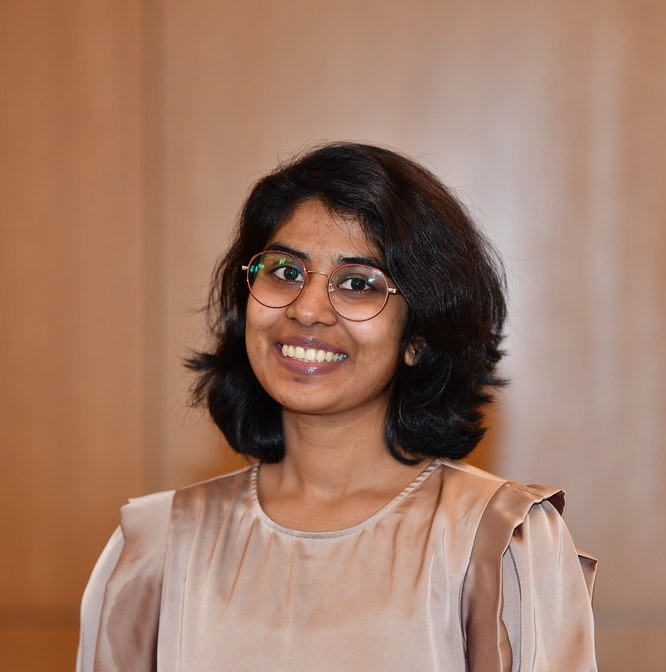}}]{Kaniz Mishty} received the B.S. degree in Electronics and Communication Engineering from Khulna University of Engineering and Technology, Bangladesh, in 2018. She is currently working towards her Ph.D. degree in ECE at Auburn University, AL, USA. Her research interests are energy efficient AI hardware design and AI/ML in CAD. She interned with Apple Inc. in Summer '22, 23 and Qualcomm in '21   on AI application in SoC and custom circuit design.
\end{IEEEbiography}

\begin{IEEEbiography}[{\includegraphics[width=\linewidth]{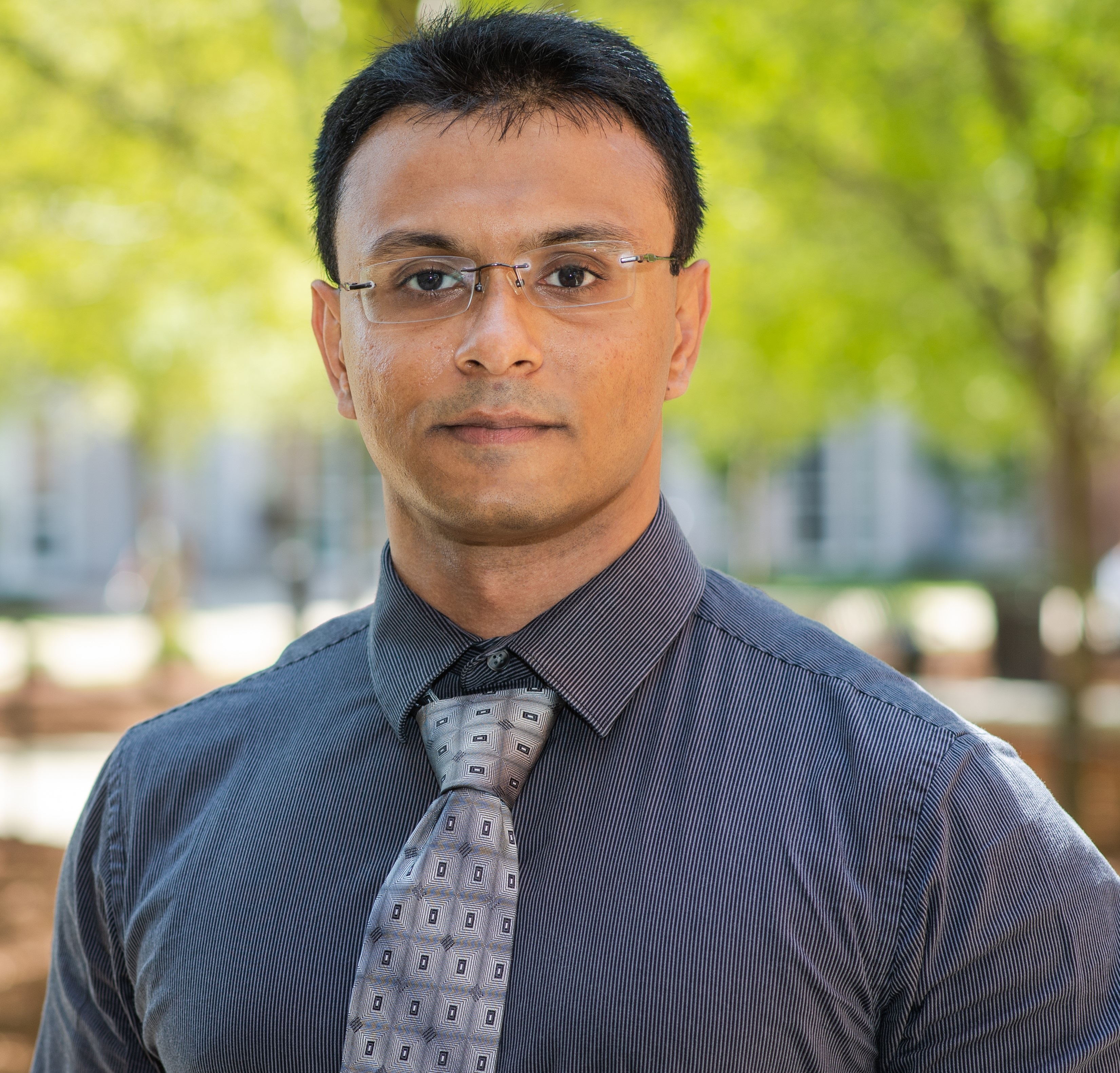}}]{Mehdi Sadi} (S'12-M'17) is an Assistant Professor at the Department of Electrical and Computer Engineering at Auburn University, Auburn, AL.  Dr. Sadi  earned his PhD in ECE from  University of Florida, Gainesville, in 2017, MS from University of California at Riverside, USA in 2011 and BS from Bangladesh University of Engineering and Technology in 2010.   Prior to joining Auburn University, he was a Senior R\&D SoC Design Engineer at Intel Corporation in Oregon. Dr. Sadi`s research focus is on developing algorithms/CAD techniques for implementation, design, reliability, and security of AI hardware. He was the recipient of SRC best in session award, Intel Xeon Design Group recognition awards, and National Science Foundation CRII award.
\end{IEEEbiography}

\begin{IEEEbiography}[{\includegraphics[width=\linewidth]{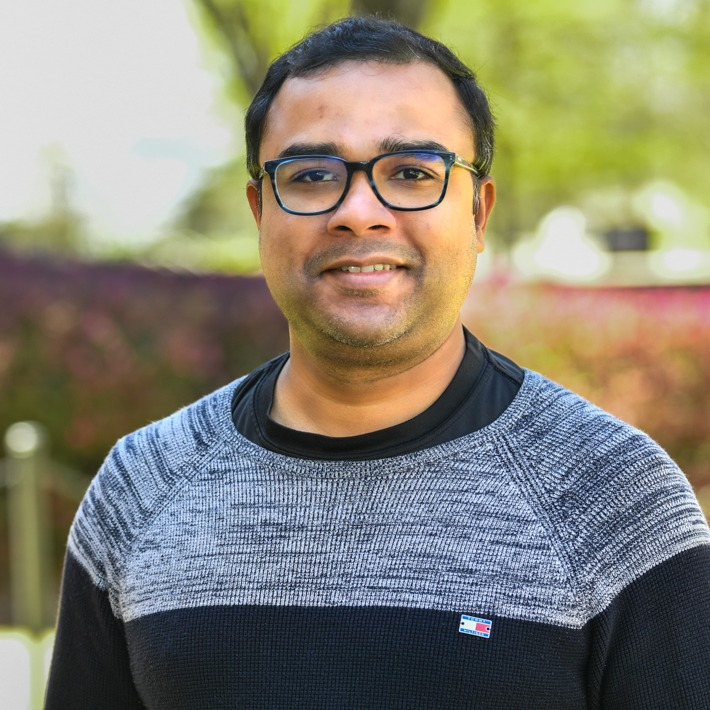}}]{Shubhra Kanti Karmaker Santu} is currently an Assistant Professor in the Department of Computer Science and Software Engineering (CSSE) at Auburn University, Alabama. With a broad interest in the field of Artificial Intelligence and Data Science, his primary research focus lies at the intersection of Natural Language Processing (NLP) and Information Retrieval (IR). He brings significant experience in NLP/IR research from both academia (Ph.D. from UIUC, Postdoc experience from MIT) and industry internships (Microsoft Research, Yahoo Research, @WalmartLabs). As a researcher, he has published 28 research papers at premier venues, including ACL, EMNLP, SIGIR, WWW, COLING, CoNLL, AACL, CIKM, IUI, ACM TIST, and ACM Computing Surveys. His research projects have been funded by multiple agencies, including the National Science Foundation (NSF), Air Force Office of Scientific Research (AFOSR), Army Research Office (ARO) and US Dept. of Agriculture (USDA). For external service, he currently serves as an action editor for the ACL Rolling Review Initiative (ARR) and as ARR's communication chair. Dr. Karmaker served as the tutorial chair for CIKM 2022. He has also served regularly as a program committee member for ACL and SIGIR-sponsored conferences for the last five years.
\end{IEEEbiography}